\documentclass{article}

\usepackage{amssymb}
\usepackage{amsthm}
\usepackage{amsmath}
\usepackage{gensymb}
\usepackage{graphicx}
\usepackage{ragged2e}
\usepackage[margin=1in]{geometry}
\usepackage{caption} 
\usepackage{afterpage}
\usepackage{mathtools}
\usepackage{algorithm}
\usepackage[noend]{algpseudocode}
\usepackage{booktabs}
\usepackage{subcaption}
\usepackage{longtable}

\makeatletter
\renewcommand{\fnum@figure}{Fig. \thefigure}
\makeatother

\begin{document}

\begin{center}
      \Large\textbf{Deep Reinforcement Learning for Multi-Agent Systems: A Review of Challenges, Solutions and Applications}\\
      \vspace{.2in}
      \large\text{Thanh Thi Nguyen, Ngoc Duy Nguyen, and Saeid Nahavandi}\\
      \vspace{.1in}
      \text{Institute for Intelligent Systems Research and Innovation}\\
      \text{Deakin University, Waurn Ponds, Victoria, Australia}\\
      %%\vspace{.1in}
      \text{E-mail: thanh.nguyen@deakin.edu.au.}\\
      \text{Tel: +61 3 52278281. Fax: +61 3 52271046.}
\end{center}
\vspace{.2in}

\begin{abstract}
Reinforcement learning (RL) algorithms have been around for decades and employed to solve various sequential decision-making problems. These algorithms however have faced great challenges when dealing with high-dimensional environments. The recent development of deep learning has enabled RL methods to drive optimal policies for sophisticated and capable agents, which can perform efficiently in these challenging environments. This paper addresses an important aspect of deep RL related to situations that require multiple agents to communicate and cooperate to solve complex tasks. A survey of different approaches to problems related to multi-agent deep RL (MADRL) is presented, including non-stationarity, partial observability, continuous state and action spaces, multi-agent training schemes, multi-agent transfer learning. The merits and demerits of the reviewed methods will be analyzed and discussed, with their corresponding applications explored. It is envisaged that this review provides insights about various MADRL methods and can lead to future development of more robust and highly useful multi-agent learning methods for solving real-world problems.
\end{abstract}

\justify
%%\begin{flushleft}

%% main text
\section{Introduction}
\label{sec1}
\begin{figure}[!t]
\centering
\includegraphics[width=0.85\columnwidth]{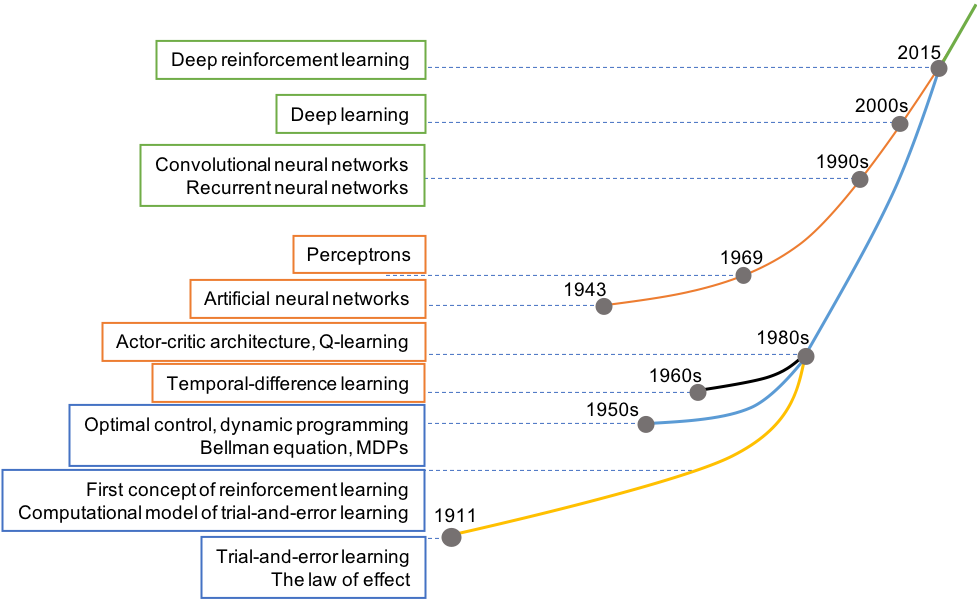}
\caption{Emergence of deep RL through different essential milestones.}
\label{fig_milestone} 
\end{figure}

Reinforcement learning was instigated by a \emph{trial and error} (TE) procedure, conducted by Thorndike in an experiment on cat's behaviors in 1898 \cite{Thorndike1898}. In 1954, Minsky \cite{Minsky1954} designed the first neural computer named Stochastic Neural-Analog Reinforcement Calculators (SNARCs), which simulated the rat's brain to solve the maze puzzle. SNARCs remarked the uplift of TE learning to a computational period. Almost two decades later, Klopf \cite{Klopf1972} integrated the mechanism of \emph{temporal-difference} (TD) learning from psychology into the computational model of TE learning. That integration succeeded in making TE learning a feasible approach to large systems. In 1989, Watkins and Dayan \cite{Watkins1992} brought the theory of optimal control \cite{Bellman1952} including \emph{Bellman equation} and \emph{Markov decision process} together with temporal-difference learning to form a well-known \emph{Q-learning}. Since then, Q-learning has been applied to solve various real-world problems, but it is unable to solve high-dimensional problems where the number of calculations increases drastically with number of inputs. This problem, known as the \emph{curse of dimensionality}, exceeds the computational constraint of conventional computers. In 2015, Mnih et al. \cite{Mnih2015} made an important breakthrough by combining \emph{deep learning} with RL to partially overcome the curse of dimensionality. Deep RL has become a normative approach in artificial intelligence, attracting significant attention from the research community since then. Milestones of the development of RL are presented in Fig. \ref{fig_milestone}, which span from the trial and error method to deep RL. 

RL originates from animal learning in psychology and thus it can mimic human learning ability to select actions that maximize long-term profit in their interactions with the environment. Therefore, RL can be used to develop an agent that is comparable to the human performance. For instance, RL has been widely used in robotics and autonomous systems, e.g. Mahadevan and Connell \cite{Mahadevan1992} designed a robot that can push cubes (1992); Schaal \cite{Schaal1997} created a humanoid robot that can effectively solve the pole-balancing task (1997); Benbrahim and Franklin \cite{Benbrahim1997} made a biped robot that can learn to walk without any knowledge of the environment (1997); Riedmiller et al. \cite{Riedmiller2009} built a soccer robot team (2009); and Muelling et al. \cite{Mulling2013} trained a robot to play table tennis (2013).

Modern RL is truly marked by the success of deep RL in 2015 when Mnih et al. \cite{Mnih2015} made use of a structure named \emph{deep Q-network} (DQN) in creating an agent that outperformed a professional player in a series of 49 classic Atari games \cite{Bellemare2013}. In 2016, Google's DeepMind created a self-taught AlphaGo program that could beat the best professional Go players, including China's Ke Jie and Korea's Lee Sadol \cite{Silver2016}. Deep RL has also been used to solve MuJuCo physic problems \cite{Duan2016} and 3D maze games \cite{Beattie2016}. In 2017, OpenAI announced a bot that could beat the best professional gamer on the online game Dota 2, which is supposed to be more complicated than the Go game. These fates provide the necessary impetus to enterprise corporations such as Google, Tesla, and Uber in their race to make self-driving cars. More importantly, deep RL has become a promising approach to solving complex real-world problems and has also been a huge contribution to the field of artificial intelligence.

RL is a research theme that distincts from other related concepts in artificial intelligence. Historically, there had been a confusion between RL and \emph{supervised learning} (SL) since the 1960s. It was not until 1981 that Sutton and Barto \cite{Sutton1998} shed a light on the discrepancy between the two learning methods. Concisely, SL is learning from data that define input and corresponding output (often called ``labelled" data) by an external supervisor, whereas RL is learning by interacting with the unknown environment. In the former case, it maybe infeasible to collect all possible behaviors in the real world to feed the algorithm. However, in the latter case, an RL agent conducts a TE procedure to gain experiences and improve itself over time. Furthermore, RL is not an \emph{unsupervised learning} (UL) method. UL is learning to explore the hidden structure of data where output information is unknown (``unlabelled" data). In contrast, RL is a goal-directed learning, i.e., it constructs a learning model that clearly specifies output to maximize the long-term profit. Finally, deep RL distinguishes with deep learning method. Deep learning uses multi-layer neural networks to learn a problem in different levels of abstraction \cite{Deng2014}. Deep RL leverages deep learning as an approximator to deal with high-dimensional data. This fact makes deep RL a promising approach to solving complex real-world problems.

As real-world problems have become increasingly complicated, there are many situations where a single deep RL agent is not able to cope with. In such situations, the applications of a \emph{multi-agent system} (MAS) are indispensable. In an MAS, agents must compete or cooperate to obtain the best overall results. Examples of such systems include multi-player online games, cooperative robots in the production factories, traffic control systems, and autonomous military systems like unmanned aerial vehicles, surveillance, and spacecraft. Among many applications of deep RL in the literature, there is a large number of studies using deep RL in MAS, henceforth \emph{multi-agent deep RL} (MADRL). Extending from a single agent domain to a multi-agent environment creates several challenges. Previous surveys considered different perspectives, for example, Busoniu et al. \cite{Busoniu2008} examined the stability and adaptation aspects of agents, Bloembergen et al. \cite{Bloembergen2015} analysed the evolutionary dynamics, Hernandez-Leal et al. \cite{Hernandez-Leal2018} considered emergent behaviors, communication and cooperation learning perspectives, and Silva et al. \cite{Silva2018} reviewed methods for knowledge reuse autonomy in \emph{multi-agent RL} (MARL). This paper presents an overview of technical challenges in multi-agent learning as well as deep RL approaches to these challenges. We cover numerous MADRL perspectives, including non-stationarity, partial observability, multi-agent training schemes, transfer learning in MAS, and continuous state and action spaces in multi-agent learning. Applications of MADRL in various fields are also reviewed and analysed in the current study. In the last section, we present extensive discussions and interesting future research directions of MADRL.

\section{Background: Reinforcement Learning}
\label{sec2}
\subsection{Preliminary}
\label{sec:2.1}

\begin{figure}[!t]
\centering
\includegraphics[width=0.45\columnwidth]{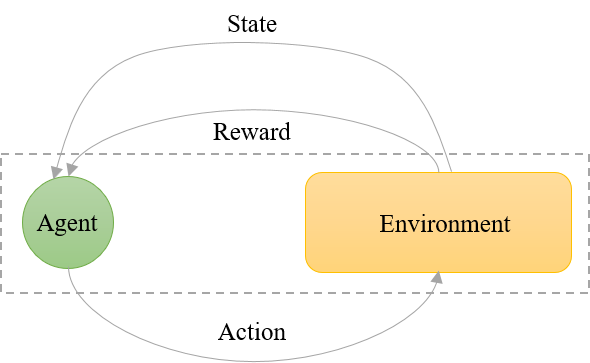}
\caption{A single agent interacting with its environment.}
\label{fig_single_agent} 
\end{figure}

\begin{figure}[!h]
\centering
\includegraphics[width=3.5in]{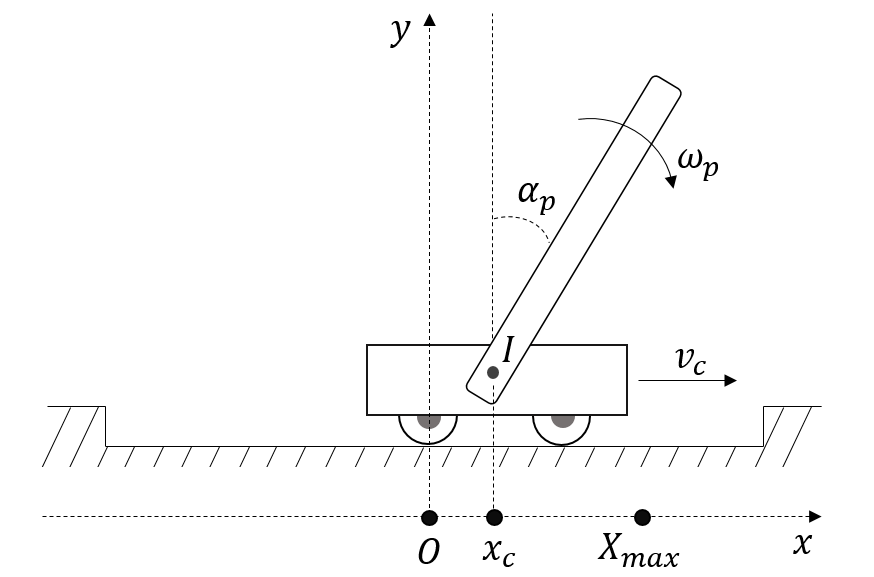}
% where an .eps filename suffix will be assumed under latex, 
% and a .pdf suffix will be assumed for pdflatex; or what has been declared
% via \DeclareGraphicsExtensions.
\caption{A 2D pole-balancing task.}
\label{fig_pole_balance} 
\end{figure}

RL is a TE learning 1) by interacting directly with the environment 2) to self-teach over time and 3) eventually achieve designating goal. Specifically, RL defines any decision maker (learner) as an \emph{agent} and anything outside the agent as an \emph{environment}. The interactions between agent and environment are described via three essential elements: state $s$, action $a$, and reward $r$, as illustrated in Fig.~\ref{fig_single_agent} \cite{Sutton1998}. The state of the environment at time-step $t$ is denoted as $s_t$. Thereby, the agent examines $s_t$ and performs a corresponding action $a_t$. The environment then alters its state $s_t$ to $s_{t+1}$ and provides a feedback reward $r_{t+1}$ to the agent. For instance, Fig.~\ref{fig_pole_balance} illustrates one of the earliest problems in RL literature, a 2D pole-balancing task. In this problem, a state of the environment at time-step $t$ can be presented by a 4-tuple $s_t$ = $[x_c, v_c, \alpha_p, \omega_p]_t$, where $x_c$ denotes \emph{x}-coordinate of the cart in Cartesian coordinate system $O_{xy}$, $v_c$ presents velocity of the cart along the track, $\alpha_p$ is the angle created by the pole and axis $O_y$, and $\omega_p$ indicates the angular velocity of the pole around center \emph{I}. The agent can produce two possible actions at each time-step $t$: exert a unit force ($|\vec{F}|=1$) to the cart along axis $O_x$ from left to right $a_t=\vec{F}$ or from right to left $a_t=-\vec{F}$. The agent is given a feedback reward $r_{t+1}=+1$ for every action that can keep the pole upright and $r_{t+1}=0$ otherwise. Therefore, the agent's goal is to keep the pole upright as long as possible and ultimately maximize the accumulated feedback reward.

Typically, the interactions between agent and environment can be presented by a series of states, actions, and rewards: $s_0$, $a_0$, $r_1$, $s_1$, $a_1, ..., r_n$, $s_n$. Although $n$ can approach to infinity, we often limit $n$ in practice by defining a terminal state $s_n=s_T$. In this case, a series of states, actions, and rewards from initial state to terminal state is called an \emph{episode}. For example, in pole-balancing task, we can define a terminal state as if $|\alpha_p| > 10\degree$ or $|x_c| > X_{max}$.

The next step is to formalize the agent's decision by defining a concept of \emph{policy}. A policy $\pi$ is a mapping function from any perceived state $s$ to action $a$ taken from that state. A policy is \emph{deterministic} if the probability of choosing an action $a$ from $s$: $p(a|s)=1$ for all state $s$. In contrast, the policy is \emph{stochastic} if there exists a state $s$ so that $p(a|s)<1$. In either case, we can define the policy $\pi$ as a probability distribution of candidate actions that will be selected from a certain state as below:

\begin{equation}
\pi = \Psi(s) = \left\lbrace\ p(a_i|s)\ \bigg\vert\ \forall a_i \in \Delta_\pi\ \wedge \sum_{i}p(a_i|s)=1\right\rbrace,
\label{eq:1}
\end{equation}

\noindent
where $\Delta_\pi$ represents all candidate actions (\emph{action space}) of policy $\pi$. For clarity, we assume that the action space is discrete because the continuous case can be straightforwardly inferred by using integral notation. Furthermore, we presume that the next state $s_{t+1}$ and feedback reward $r_{t+1}$ are entirely determined by the current state-action pair $(s_t,a_t)$ regardless of the history. Any RL problem satisfies this ``memoryless" condition is known as \emph{Markov decision process} (MDP). Therefore, the \emph{dynamics} (model) of an RL problem is completely specified by giving all \emph{transition probabilities} $p(a_i|s)$. Based on (\ref{eq:1}), we can define a deterministic policy $\pi_d$:

\begin{equation*}
    \pi_d = \Psi_d(s) =
    \begin{cases}
      1, & a_i = a(s) \wedge a(s) \in \Delta_{\pi_d}\\
      0, & \forall a_i \in \Delta_{\pi_d} \wedge a_i \neq a(s)
    \end{cases},
\label{eq:2}
\end{equation*}

\noindent
where $a(s)$ denotes the designating action taken at state $s$. Deterministic policy is desirable in practical application because it has a predictable behavior, which is a crucial factor for designing an effective RL algorithm. In practice, we can derive a deterministic policy $\pi_d$ from a stochastic policy $\pi$ using the following rule:

\begin{equation}
    \boldsymbol{R_1}:\pi \mapsto \pi_d = \Psi_d(s) =
    \begin{cases}
      1, & a_i = a_j \wedge j = \underset{k}{\arg\max}\ \pi(s,a_k) \\
      0, & \forall a_i \in \Delta_{\pi} \wedge a_i \neq a_j
    \end{cases},
\label{eq:3}
\end{equation}

\noindent
where $\pi(s,a_k)$ denotes the probability of taking action $a_k \in \Delta_\pi$ in state $s$ using policy $\pi$ and $\Delta_{\pi_d} = \Delta_{\pi}$.

Initially, the agent is assigned a random policy $\pi_0$. It adjusts the policy $\pi_0$ to improve itself by interacting with the environment in a TE learning manner. In this respect, we call that policy $\pi_{t+1}$ is better than policy $\pi_{t}$ and denoted as $\pi_{t+1} > \pi_t$. Therefore, we have a series of policies improved over time as follows:

\begin{equation*}
\pi_0 < \pi_1 < ... < \pi_t < \pi_{t+1} < ... < \pi^*.
\label{eq:4}
\end{equation*}

\noindent
This process, named \emph{policy improvement}, is repeated until the agent cannot find any policy better than the \emph{optimal policy} $\pi^*$. By this definition, however, we still do not know exactly how to compare two policies and decide which one is better. In the next subsection, we will review other metrics that can be used to evaluate a policy and then we can use these metrics to compare how ``good" between different policies.

\subsection{Bellman equation}
\label{sec:2.2}

Remind that the agent receives a feedback reward $r_{t+1}$ for every time-step $t$ until it reaches the terminal state $s_T$. However, the immediate reward $r_{t+1}$ does not represent the long-term profit, we instead leverage a generalized \emph{return value} $R_t$ at time-step $t$:

\begin{equation}
R_t = r_{t+1} + \gamma r_{t+2} + \gamma^2 r_{t+3} + ... + \gamma^{T-t-1} r_T = \sum_{i=0}^{T-t-1}\gamma^i r_{t+i+1},
\label{eq:5}
\end{equation}

\noindent
where $\gamma$ is a discounted factor so that $0 \leq \gamma < 1$. The agent becomes farsighted when $\gamma$ approaches to $1$ and vice versa the agent becomes shortsighted when $\gamma$ is close to $0$. Apparently, we often select $\gamma$ closing to $1$ in practice.

The next step is to define a \emph{value function} that is used to evaluate how ``good" of a certain state $s$ or a certain state-action pair $(s,a)$. Specifically, the value function of state $s$ under policy $\pi$ is calculated by obtaining expected return value from $s$ \cite{Sutton1998}: $V_\pi(s) = \mathop{\mathbb{E}}[R_t | s_t = s, \pi]$. Likewise, the value function of state-action pair $(s,a)$ is $Q_\pi(s,a) = \mathop{\mathbb{E}}[R_t | s_t=s,a_t=a,\pi]$. We can leverage value functions to compare how ``good" between two policies $\pi$ and $\pi'$ using the following rule \cite{Sutton1998}:

\begin{equation}
\pi \leq \pi' \Longleftrightarrow \bigg[V_\pi(s) \leq V_{\pi'}(s)\ \forall s\bigg] \vee \bigg[Q_\pi(s,a) \leq Q_{\pi'}(s,a)\ \forall (s,a)\bigg].
\label{eq:6}
\end{equation}

Based on (\ref{eq:5}), we can expand $V_\pi(s)$ and $Q_\pi(s,a)$ to present the relationship between two consecutive states $s=s_t$ and $s'=s_{t+1}$ as below \cite{Sutton1998}:

\begin{equation}
V_\pi(s) = \sum_{a}\pi(s,a)\sum_{s'}p(s'|s,a)\bigg(\mathop{}\mathbb{W}_{s\rightarrow s'|a} + \gamma V_\pi(s')\bigg),
\label{eq:7}
\end{equation}

\noindent
and

\begin{equation}
Q_\pi(s,a) = \sum_{s'}p(s'|s,a)\bigg(\mathop{}\mathbb{W}_{s\rightarrow s'|a} + \gamma V_\pi(s')\bigg),
\label{eq:8}
\end{equation}

\begin{figure}[!t]
\centering
\includegraphics[width=5.in]{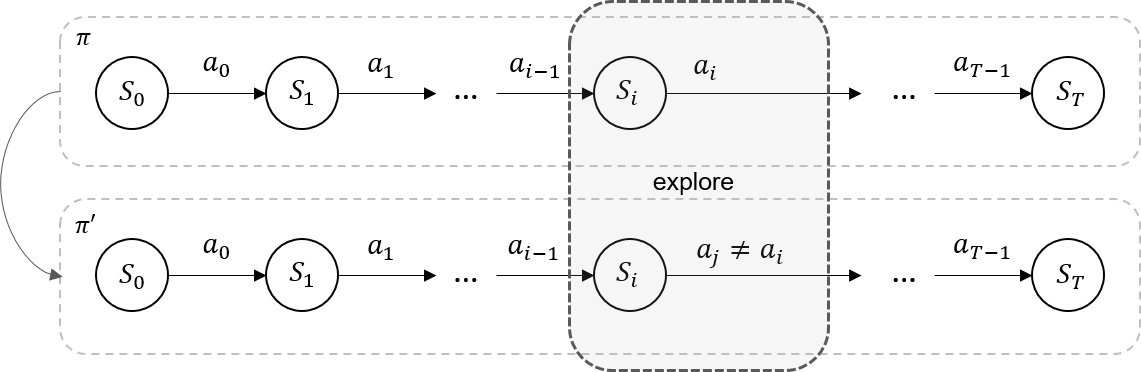}
% where an .eps filename suffix will be assumed under latex, 
% and a .pdf suffix will be assumed for pdflatex; or what has been declared
% via \DeclareGraphicsExtensions.
\caption{A naive approach to finding a better policy $\pi'$ from $\pi$.}
\label{fig_better_policy} 
\end{figure}

\noindent
where $\mathop{\mathbb{W}}_{s\rightarrow s'|a} = \mathop{\mathbb{E}}[r_{t+1} | s_t = s, a_t = a, s_{t+1} = s']$. Solving (\ref{eq:7}) or (\ref{eq:8}), we can find value function $V(s)$ or $Q(s,a)$, respectively. Equations (\ref{eq:7}) and (\ref{eq:8}) are called Bellman equations and widely used in policy improvement. For example, Fig.~\ref{fig_better_policy} describes a naive approach to improve a deterministic policy $\pi$. By using a TE approach, we can ``explore" a different action $a_j \neq a_i$ taken at state $s_i$ so that $Q_\pi(s_i,a_j) > Q_\pi(s_i,a_i)$. We include $a_j$ as a new action taken at $s_i$ in a derived policy $\pi'$ while keeping other pairs of state-action unchanged. Based on (\ref{eq:6}) and (\ref{eq:8}), we can infer that $\pi < \pi'$. Therefore, it is straightforward to select a ``greedy" action $a_j$ so that $Q_\pi(s_i,a_j)$ attains maximum values. Finally, we can derive an improved policy $\pi'$ from $\pi$ using the following rule:

\begin{equation}
    \boldsymbol{R_2}: \pi \mapsto \pi' = \Psi'(s) =
    \begin{cases}
      1, & a_i = a_j \wedge j = \underset{k}{\arg\max}\ Q_{\pi}(s,a_k) \\
      0, & \forall a_i \in \Delta_{\pi} \wedge a_i \neq a_j
    \end{cases}.
\label{eq:9}
\end{equation}

\noindent
This process is iterated for all pairs of $(s_i,a_i)$ until we find an optimal solution $\pi^*$. The idea indeed can be generalized to any stochastic policy $\pi$. 

Instead of repeating policy improvement process, we can estimate directly the value function of optimal policy $\pi^*$ using the following \emph{optimality Bellman equation} \cite{Sutton1998}:

\begin{equation}
V_{\pi^*}(s) = \max_{a}\sum_{s'}p(s'|s,a)\bigg(\mathop{}\mathbb{W}_{s\rightarrow s'|a} + \gamma V_{\pi^*}(s')\bigg),
\label{eq:10}
\end{equation}

\noindent
and

\begin{equation}
Q_{\pi^*}(s,a) = \sum_{s'}p(s'|s,a)\bigg(\mathop{}\mathbb{W}_{s\rightarrow s'|a} + \gamma \max_{a'}Q_{\pi^*}(s',a')\bigg).
\label{eq:11}
\end{equation}

\noindent
After solving optimality Bellman equations (\ref{eq:10}) or (\ref{eq:11}), we can derive an optimal deterministic policy $\pi^*$ using (\ref{eq:9}).

Although we can use \emph{dynamic programming} to approximate the solutions of Bellman equations, it requires the complete dynamics information of the problem. Therefore, it is infeasible due to the lack of memory and computational power of conventional computer when the number of states is large. In the next subsection, we will review two \emph{model-free} RL methods (require no knowledge of transition probabilities $p(a_i|s)$) to approximate the value functions.

\subsection{RL methods}
\label{sec:2.3}
In this section, we review two well-known learning schemes in RL: Monte-Carlo and temporal-difference learning. These methods do not require the dynamics information of the environment, i.e., they can deal with larger state-space problems than trivial dynamic programming approaches.

\subsubsection{Monte-Carlo method}
\emph{Monte-Carlo} (MC) method estimates value function by repeatedly generating episodes and recording average return at each state or each state-action pair. Therefore, the state-value function is calculated as follows: 

\begin{equation*}
V_{\pi}^{MC}(s) = \lim_{i \to +\infty}\mathop{\mathbb{E}}\bigg[r^i(s_t)\ \big|\ s_t = s, \pi\bigg],
\label{eq:12}
\end{equation*}

\noindent
where $r^i(s_t)$ denotes observed return at state $s_t$ in episode $i$th. Similarly, we have value function of state-action pair:

\begin{equation*}
Q_{\pi}^{MC}(s,a) = \lim_{i \to +\infty}\mathop{\mathbb{E}}\bigg[r^i(s_t,a_t)\ \big|\ s_t = s, a_t = a, \pi\bigg].
\label{eq:13}
\end{equation*}

\noindent
MC method does not require any knowledge of transition probabilities, i.e., MC method is model-free. However, this approach has made two essential assumptions to ensure the convergence happens: 1) the number of episodes is large and 2) every state and every action must be visited with a significant number of times. To make this ``exploration" possible, we often use \emph{$\epsilon$-greedy} strategy in policy improvement instead of (\ref{eq:9}):

\begin{equation}
    \boldsymbol{R_3}:\pi \mapsto \pi' = \Psi'(s) =
    \begin{cases}
      1-\epsilon + \frac{\epsilon}{|\Delta_{\pi}(s)|}, & a_i = a_j \wedge j = \underset{k}{\arg\max}\ Q_{\pi}(s,a_k) \\
      \frac{\epsilon}{|\Delta_{\pi}(s)|}, & \forall a_i \in \Delta_{\pi} \wedge a_i \neq a_j
    \end{cases},
\label{eq:14}
\end{equation}

\noindent
where $|\Delta_{\pi}(s)|$ denotes number of candidate actions taken in state $s$ and $0 < \epsilon < 1$. Generally, MC algorithms are divided into two groups: \emph{on-policy} and \emph{off-policy}. In on-policy methods, we use policy $\pi$ for both evaluation and exploration purpose. Therefore, the policy $\pi$ must be stochastic or \emph{soft}. In contrast, off-policy uses different policy $\pi' \neq \pi$ to generate the episodes and hence $\pi$ can be deterministic. Although off-policy is desirable due to its simplicity, on-policy method is more stable when working with continuous state-space problems and when using together with a function approximator (such as neural networks) \cite{Tsitsiklis1997}. 

\subsubsection{Temporal-difference method}
Similar to MC, \emph{temporal-difference} (TD) learning is also learning from experiences (model-free method). However, unlike MC, TD learning does not wait until the end of episode to make an update. It makes an update on every step within the episode by leveraging 1-step Bellman equation (\ref{eq:7}) and hence possibly providing faster convergence:

\begin{equation*}
\boldsymbol{U_1}: V^{i}(s_t) \longleftarrow \alpha V^{i-1}(s_t) + (1 - \alpha)\bigg(r_{t+1} + \gamma V^{i-1}(s_{t+1})\bigg),
\label{eq:15}
\end{equation*}

\noindent
where $\alpha$ is step-size parameter and $0 < \alpha < 1$. TD learning uses previous estimated values $V^{i-1}$ to update the current ones $V^{i}$, which is known as \emph{bootstrapping} method. Basically, bootstrapping method learns faster than non-bootstrapping ones in most of the cases \cite{Sutton1998}. TD learning is also divided into two categories: on-policy TD control (\emph{Sarsa}) and off-policy TD control (\emph{Q-learning}). In Sarsa, the algorithm estimates value function of state-action pair based on (\ref{eq:8}):

\begin{equation*}
\boldsymbol{U_2}:Q^{i}(s_t,a_t) \longleftarrow \alpha Q^{i-1}(s_t,a_t) + (1 - \alpha)\bigg(r_{t+1} + \gamma Q^{i-1}(s_{t+1},a_{t+1})\bigg).
\label{eq:16}
\end{equation*}

On the other hand, Q-learning uses 1-step optimality Bellman equation (\ref{eq:11}) to perform the update, i.e., Q-learning directly approximates value function of optimal policy:

\begin{equation}
\boldsymbol{U_3}:Q^{i}(s_t,a_t) \longleftarrow \alpha Q^{i-1}(s_t,a_t) + (1 - \alpha)\bigg(r_{t+1} + \gamma \overbrace{\max_{a_{t+1}^j} Q^{i-1}(s_{t+1},a_{t+1}^j)}^{\text{deterministic subpolicy}}\bigg).
\label{eq:17}
\end{equation}

\begin{table}[!t]
\centering
\caption{Characteristics of RL methods}
\tiny
\resizebox{\textwidth}{!}{%
\begin{tabular}{ccc}
\hline
\hline
Category & Pros & Cons \\
\hline
Model-free & $\bullet$ Dynamics of environment & $\circ$ Requires ``exploration"\\
           &is unknown&condition\\
           &$\bullet$ Deal with larger state-space&\\
           &environments&\\
\hline
On-policy &$\bullet$ Stable when using&$\circ$ Policy must be stochastic\\
          &with function approximator&\\
          &$\bullet$ Suitable with continuous&\\
          &state-space problems&\\
\hline
Off-policy &$\bullet$ Simplify algorithm design&$\circ$ Unstable when using\\
           &$\bullet$ Can tackle with&with function approximator\\
           &different kinds of problems&\\
           &$\bullet$ Policy can be deterministic&\\
\hline
Bootstrapping &$\bullet$ Learn faster in most cases& $\circ$ Not as good as nonbootstrapping\\
          &&methods on mean square error\\
\hline
\hline
\end{tabular}}
\label{table:1}
\end{table}

\noindent
We notice that the operator $max$ in update rule (\ref{eq:17}) substitutes for a deterministic policy. This strongly explains why Q-learning is off-policy.

In practice, MC and TD learning often use table memory structure (\emph{tabular method}) to save value function of each state or each state-action pair. This makes them inefficient due to lack of memory in solving complicated problems where number of states is large. Therefore, \emph{actor-critic} (AC) architecture is designed to subdue this limitation. Specifically, AC includes two separate memory structures for an agent: \emph{actor} and \emph{critic}. Actor structure is used to select a suitable action according to the observed state and transfer to critic structure for evaluation. Critic structure uses the following TD error to decide future tendency of the selected action:

\begin{equation*}
\delta(a_t) = \beta \bigg(r_{t+1} + \gamma V(s_{t+1})\bigg) - (1-\beta) V(s_t),
\label{eq:18}
\end{equation*}

\noindent
where $0 < \beta < 1$; and if $\delta(a_t) > 0$, the tendency to select the action $a_t$ in the future is high and vice versa. Furthermore, AC can be on-policy or off-policy depending on the implementation details. Table \ref{table:1} and Table \ref{table:2} summarize characteristics of RL methods and the comparisons between different RL methods, respectively.

\begin{table}[htbp]
\caption{Comparisons between RL methods}
\tiny
\centering
\resizebox{\textwidth}{!}{%
\begin{tabular}{|c|cccccc|}
\hline
Category & Dynamic & On-policy & Off-policy & Sarsa & Q-learning & AC\\
		 & programming & MC & MC & & &\\
\hline
Model-free&& \checkmark & \checkmark & \checkmark & \checkmark & \checkmark \\
On-policy&& \checkmark && \checkmark & & \checkmark \\
Off-policy&&& \checkmark && \checkmark & \checkmark \\
Bootstrapping& \checkmark &&& \checkmark & \checkmark & \checkmark \\
\hline

\hline
\end{tabular}}
\label{table:2}
\end{table}

\section{Deep RL: Single Agent}
\label{sec:3}

\subsection{Deep Q-Network}
\label{sec:3.1}

\begin{figure}[!t]
\centering
\includegraphics[width=0.9\columnwidth]{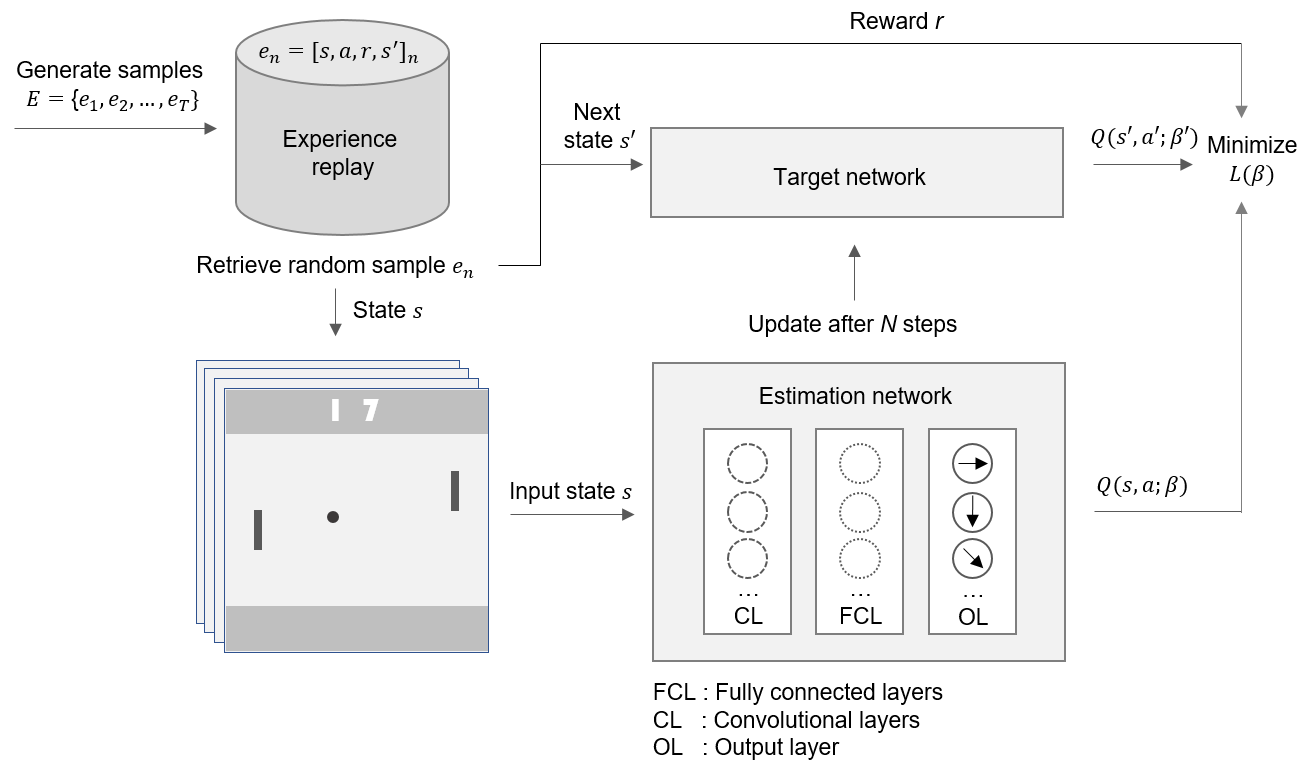}
\caption{Deep Q-network structure.}
\label{fig_DQN} 
\end{figure} 

Deep RL is a broad term that indicates the combination between deep learning and RL to deal with high-dimensional environments \cite{Arulkumaran2017, Li2017, Nguyen2017}. In 2015, Mnih et al. \cite{Mnih2015} at the first time announced the success of this combination by creating an autonomous agent that can play competently a series of 49 Atari games. Concisely, the authors proposed a novel structure named \emph{deep Q-network} (DQN) that leverages the \emph{convolutional neural network} (CNN) \cite{Krizhevsky2012} to directly interpret graphical representation of input state $s$ from the environment. The output of DQN produces Q-values of all possible actions $a \in \Delta_{\tau}$ taken at state $s$, where $\Delta_{\tau}$ denotes action space \cite{Nguyen2018a}. Therefore, DQN can be seen as a policy network $\tau$, parameterized by $\beta$, which is continually trained so as to approximate optimal policy. Mathematically, DQN uses Bellman equation to minimize the loss function $\mathcal{L}(\beta)$ as below:

\begin{equation}
\mathcal{L}(\beta) = \mathbb{E}\left[{\left(r + \gamma \max_{a'}Q(s',a'\vert\beta)-Q(s,a\vert\beta) \right)}^2 \right].
\label{eq:19}
\end{equation}

However, using neural network to approximate value function is proved to be unstable and may result in divergence due to the bias originated from correlative samples \cite{Tsitsiklis1997}. To make the samples uncorrelated, Mnih et al. \cite{Mnih2015} created a \emph{target network} $\tau'$, parameterized by $\beta'$, which is updated in every $N$ steps from estimation network $\tau$. Moreover, generated samples are stored in an \emph{experience replay memory}. Samples are then retrieved randomly from the experience replay and fed into the training process, as described in Fig.~\ref{fig_DQN}. Therefore, equation (\ref{eq:19}) can be rewritten as: 

\begin{equation}
\begin{cases}
	\mathcal{L}_{\text{DQN}}(\beta) = \mathbb{E}\left[{\left(r + \gamma \max_{a'}Q(s',a'\vert\beta')-Q(s,a\vert\beta) \right)}^2 \right] \\
	\beta' \longleftarrow \beta\ \text{for every N steps}
\end{cases}
\label{eq:20}
\end{equation}

Although DQN basically solved a challenging problem in RL, the curse of dimensionality, this is just a rudimental step in solving completely real-world applications. DQN has numerous drawbacks, which can be remedied by different schemes, from a simple form to complicated modifications that we will discuss in the next subsection.

\subsection{DQN variants}
\label{sec:3.1}

The first and simplest form of DQN's variant is \emph{double deep Q-network} (DDQN) proposed by Hasselt \cite{Hasselt2010, Hasselt2016}. The idea of DDQN is to separate the selection of ``greedy" action from action evaluation. In this way, DDQN expects to reduce the overestimation of Q-values in the training process. In other words, the $max$ operator in equation (\ref{eq:20}) is decoupled into two different operators, as represented by the following loss function:
\begin{equation}
\mathcal{L}_{\text{DDQN}}(\beta) = \mathbb{E}\left[{\left(r + \gamma Q(s', \arg\max_{a'}Q(s',a'\vert\beta)\vert\beta')-Q(s,a\vert\beta) \right)}^2 \right]. 
\label{eq:21}
\end{equation}

\noindent
Empirical experiment results on 57 Atari games show that the normalized performance of DDQN without tuning is two times greater than DQN and three times greater than DQN when tuning.

Secondly, experience replay in DQN plays an important role to break the correlations between samples, and at the same time, remind ``rare" samples that the policy network may rapidly forget. However, the fact that selecting randomly samples from experience replay does not completely separate sample data. Specifically, we prefer rare and goal-related samples  to appear more frequent than redundancy ones. Therefore, Schaul et al. \cite{Schaul2016} proposed a \emph{prioritized experience replay} that gives priority to a sample $i$ based on its absolute value of TD error:

\begin{equation}
p_i = |\delta_i| =  |r_{i} + \gamma \max_{a} Q(s_{i}, a|\beta') - Q(s_{i-1},a_{i-1}|\beta)|
\label{eq:22}
\end{equation}

\noindent
Prioritized experience replay when combining with DDQN provides stable convergence of policy network and achieves a performance up to five times greater than DQN in terms of normalized mean score on 57 Atari games.

DQN's policy evaluation process is struggle to work in ``redundant" situations, i.e., there are more than two candidate actions that can be selected without getting any negative results. For instance, when driving a car and there are no obstables ahead, we can follow either the left lane or the right lane. If there is an obstacle ahead in the left lane, we must be in the right lane to avoid crashing. Therefore, it is more efficient if we focus only on the road and obstacles ahead. To resolve such situations, Wang et al. \cite{Wang2016} proposed a novel network architecture named \emph{dueling network}. In dueling architecture, there are two collateral networks that coexist: one network, parameterized by $\theta$, estimates state-value function $V(s|\theta)$ and the other one, parameterized by $\theta'$, estimates advantage action function $A(s,a|\theta')$. The two networks are then aggregated using the following equation to approximate Q-value function:

\begin{equation*}
Q(s,a|\theta,\theta') = V(s|\theta) + \left(A(s,a|\theta') - \frac{1}{|\Delta_\pi|}\sum_{a'}A(s,a'|\theta')\right).
\label{eq:23}
\end{equation*}

\noindent
Because dueling network represents action-value function, it was combined with DDQN and prioritized experience replay to boost the performance of the agent up to six times more than standard DQN on the Atari domain \cite{Wang2016}.

Another drawback of DQN is that it uses a history of four frames as an input to policy network. DQN is therefore inefficient to solve problems where the current state depends on a significant amount of history information such as Double Dunk or Frostbite. These games are often called partially observable MDP problems. The straightforward solution is to replace the fully-connected layer right after the last convolutional layer of the policy network with a recurrent \emph{long short term memory}, as described in \cite{Hausknecht2015}. This DQN's variant named \emph{deep recurrent Q-network} (DRQN) outperforms standard DQN up to 700 percent in games Double Dunk and Frostbite. Furthermore, Lample and Chaplot \cite{Lample2017} successfully created an agent that beats an average player on Doom, a 3D FPS (first-person shooter) environment by adding a game feature layer in DRQN. Another interesting variant of DRQN is \emph{deep attention recurrent Q-network} (DARQN) \cite{Sorokin2015}. In that paper, Sorokin et al. add attention mechanism into DRQN so that the network can focus only on important regions in the game, allowing smaller network's parameters and hence speeding the training process. As a result, DARQN achieves a score of 7263 compared with 1284 and 1421 of DQN and DRQN on game Seaquest, respectively.

\section{Deep RL: Multi-Agent}
\label{sec:4}
\begin{figure}[!t]
\centering
\includegraphics[width=0.5\columnwidth]{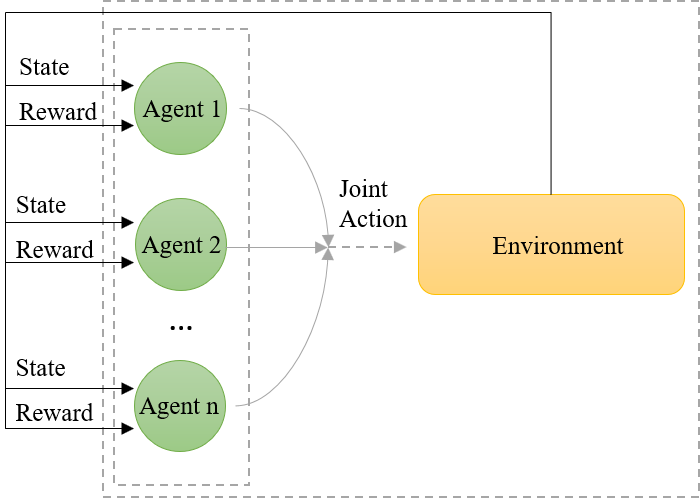}
\caption{Multiple agents interacting with the same environment.}
\label{fig_multi_agent} 
\end{figure} 

MASs have attracted great attention because they are able to solve complex tasks through the cooperation of individual agents. Within a MAS, agents communicate with each other and interact with the environment (Fig. \ref{fig_multi_agent}). In a multi-agent learning domain, the MDP is generalized to a stochastic game, or a Markov game. Let denote $n$ as the number of agents, $S$ as a discrete set of environmental states, and $A_i, i=1,2,...,n$ as a set of actions for each agent. The  joint action set for all agents is defined by $A=A_{1} \times A_{2} \times ....\times A_{n}$. The state transition probability function is represented by $p:S\times A \times S \rightarrow  [0, 1]$ and the reward function is specified as $r:S\times A \times S \rightarrow  \mathbb{R}^{n}$. The value function of each agent is dependent on the joint action and joint policy, which is characterized by $ V^{\pi}: S \times A \rightarrow \mathbb{R}^{n}$. The following subsections describe challenges and MADRL solutions as well as their applications to solve real-world problems.

\subsection{MADRL: Challenges and Solutions}
\subsubsection{Non-stationarity}
Controlling multiple agents poses several additional challenges as compared to single agent setting such as the \emph{heterogeneity} of agents, how to define suitable collective goals or the scalability to large number of agents that requires design of compact representations, and more importantly the non-stationarity problem. In a single agent environment, an agent is concerning only the outcome of its own actions. In a multi-agent domain, an agent observes not only the outcomes of its own action but also the behavior of other agents. Learning among the agents is complex because all agents potentially interact with each other and learn concurrently. The interactions among multiple agents constantly reshape the environment and lead to non-stationarity. In this case, learning among the agents sometimes causes changes in the policy of an agent, and can affect the optimal policy of other agents. The estimated potential rewards of an action would be inaccurate and therefore, good policies at a given point in the multi-agent setting could not remain so in the future. The convergence theory of Q-learning applied in single agent setting is not guaranteed to most multi-agent problems as the Markov property does not hold anymore in the non-stationary environment \cite{Hernandez-Leal2017}. Therefore, collecting and processing information must be performed with certain recurrence while ensuring that it does not affect the agents' stability. The exploration-exploitation dilemma could be more involved under multi-agent settings.

The popular \emph{independent Q-learning} \cite{Tan1993} or experience replay based DQN \cite{Mnih2015} was not designed for the non-stationary environments. Castaneda \cite{Castaneda2016} proposed two variants of DQN, namely deep repeated update Q-network (DRUQN) and deep loosely coupled Q-network (DLCQN), to deal with the non-stationarity problem in MAS. The DRUQN is developed based on the repeated update Q-learning (RUQL) model introduced in \cite{Abdallah2013, Abdallah2016}. It aims to avoid policy bias by updating the action value inversely proportional to the likelihood of selecting an action. On the other hand, DLCQN relies on the loosely coupled Q-learning proposed in \cite{Yu2015}, which specifies and adjusts an independence degree for each agent using its negative rewards and observations. Through this independence degree, the agent learns to decide whether it needs to act independently or cooperate with other agents in different circumstances. Likewise, Diallo et al. \cite{Diallo2017} extended DQN to a \emph{multi-agent concurrent DQN} and demonstrated that this method can converge in a non-stationary environment. Foerster et al. \cite{Foerster2017} alternatively introduced two methods for stabilising experience replay of DQN in MADRL. The first method uses the importance sampling approach to naturally decay obsolete data whilst the second method disambiguates the age of the samples retrieved from the replay memory using a fingerprint.

\begin{figure}[!t]
\centering
\includegraphics[width=0.85\columnwidth]{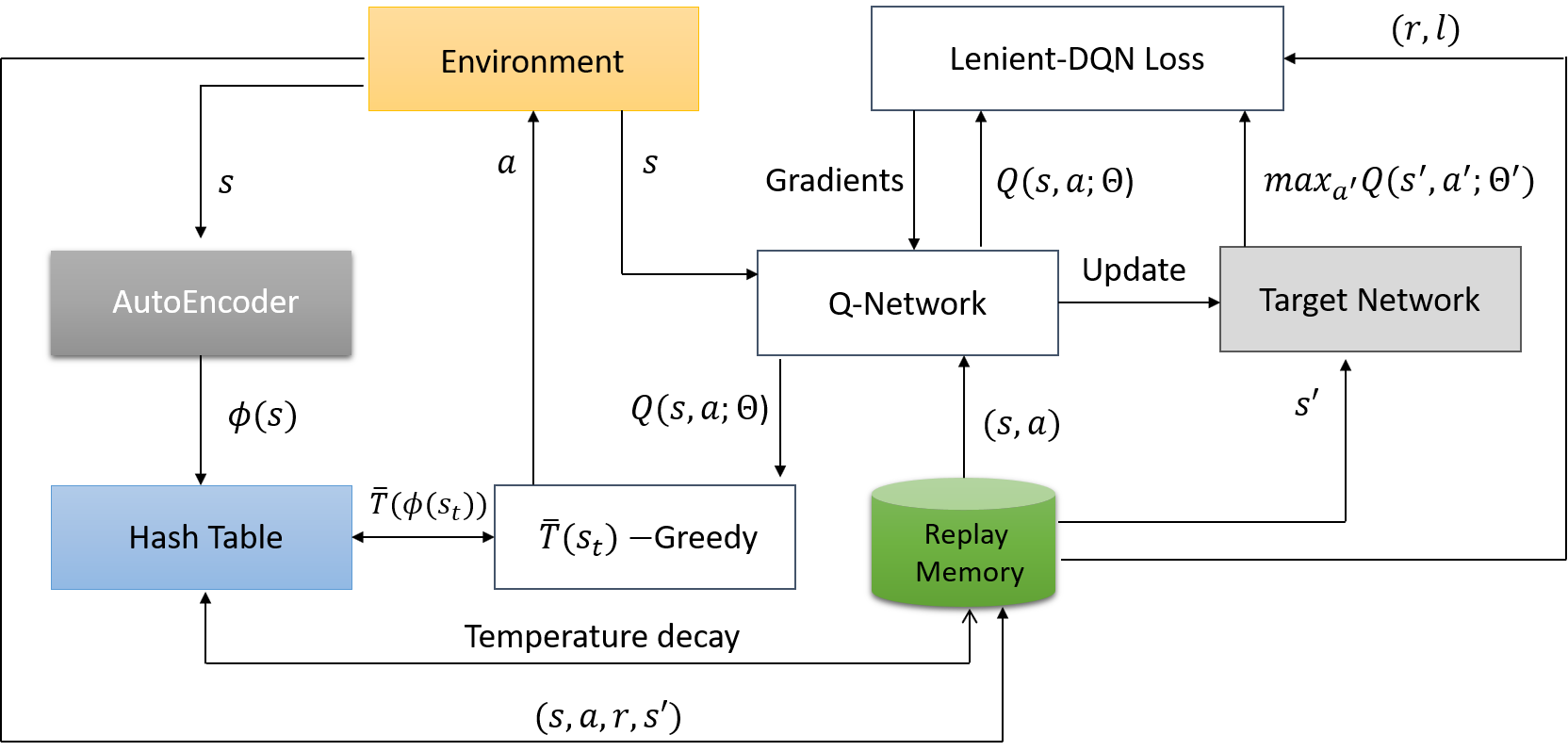}
\caption{Architecture of LDQN.}
\label{fig_LDQN} 
\end{figure}

Recently, to deal with non-stationarity due to concurrent learning of multiple agents in MAS, Palmer et al. \cite{Palmer2018} presented a method namely \emph{lenient-DQN} (LDQN) that applies leniency with decaying temperature values to adjust policy updates sampled from the experience replay memory (Fig. \ref{fig_LDQN}). That method is applied to the coordinated multi-agent object transportation problems and its performance is compared with the \emph{hysteretic-DQN} (HDQN) \cite{Omidshafiei2017}. The experimental results demonstrate the superiority of LDQN against HDQN in terms of convergence to optimal policies in a stochastic reward environment. The notion of leniency along with a scheduled replay strategy were also incorporated into the \emph{weighted double deep Q-network} (WDDQN) in \cite{Zheng2018} to deal with non-stationarity in MAS. Experiments show the better performance of WDDQN against double DQN in two multi-agent environments with stochastic rewards and large state space. 

\subsubsection{Partial observability}
In real-world applications, there are many circumstances where agents only have partial observability of the environment. In other words, complete information of states pertaining to the environment is not known to the agents when they interact with the environment. In such situations, the agents observe partial information about the environment, and need to make the “best” decision during each time step. This type of problem can be modelled using the \emph{partially observable Markov decision process} (POMDP).

In the current literature, a number of deep RL models have been proposed to handle POMDP. Hausknecht and Stone \cite{Hausknecht2015} proposed \emph{deep recurrent Q-network} (DRQN) based on a long short term memory network. With the recurrent structure, the DRQN-based agents are able to learn the improved policy in a robust sense in the partially observable environment. Unlike DQN, DRQN approximates $Q(o,a)$, which is a Q-function with observation $o$ and action $a$, by a recurrent neural network. DRQN treats a hidden state of the network $h_{t-1}$ as an internal state. The DRQN therefore is characterized by the Q-function $(o_t,h_{t-1},a;\theta_i)$ where $\theta_i$ is the parameters of the network at the $i$th training step. In \cite{Foerster2016b}, DRQN is extended to \emph{deep distributed recurrent Q-network} (DDRQN) to handle multi-agent POMDP problems. The success of DDRQN is relied on three notable features, i.e. last-action inputs, inter-agent weight sharing, and disabling experience replay. The first feature, i.e. last-action inputs, requires the provision of previous action of each agent as input to its next step. The inter-agent weight sharing means that all agents use weights of only one network, which is learned during the training process. The disabling experience replay simply excludes the experience replay feature of DQN. DDRQN therefore learns a Q-function of the form $Q(o_t^m,h_{t-1}^m,m,a_{t-1}^m,a_t^m;\theta_i)$ where each agent receives its own index $m$ as input. Weight sharing decreases learning time because it reduces the number of parameters to be learned. Although each agent has different observation and hidden state, this approach however assumes that agents have the same set of actions. To address complicated problems, autonomous agents often have different sets of actions. For example, UAVs manoeuvre in the air whilst robots operate on the ground. Therefore, action spaces of UAVs and robots are different and thus the inter-agent weight sharing feature cannot be applied. 

\begin{figure}[!t]
\centering
\includegraphics[width=0.75\columnwidth]{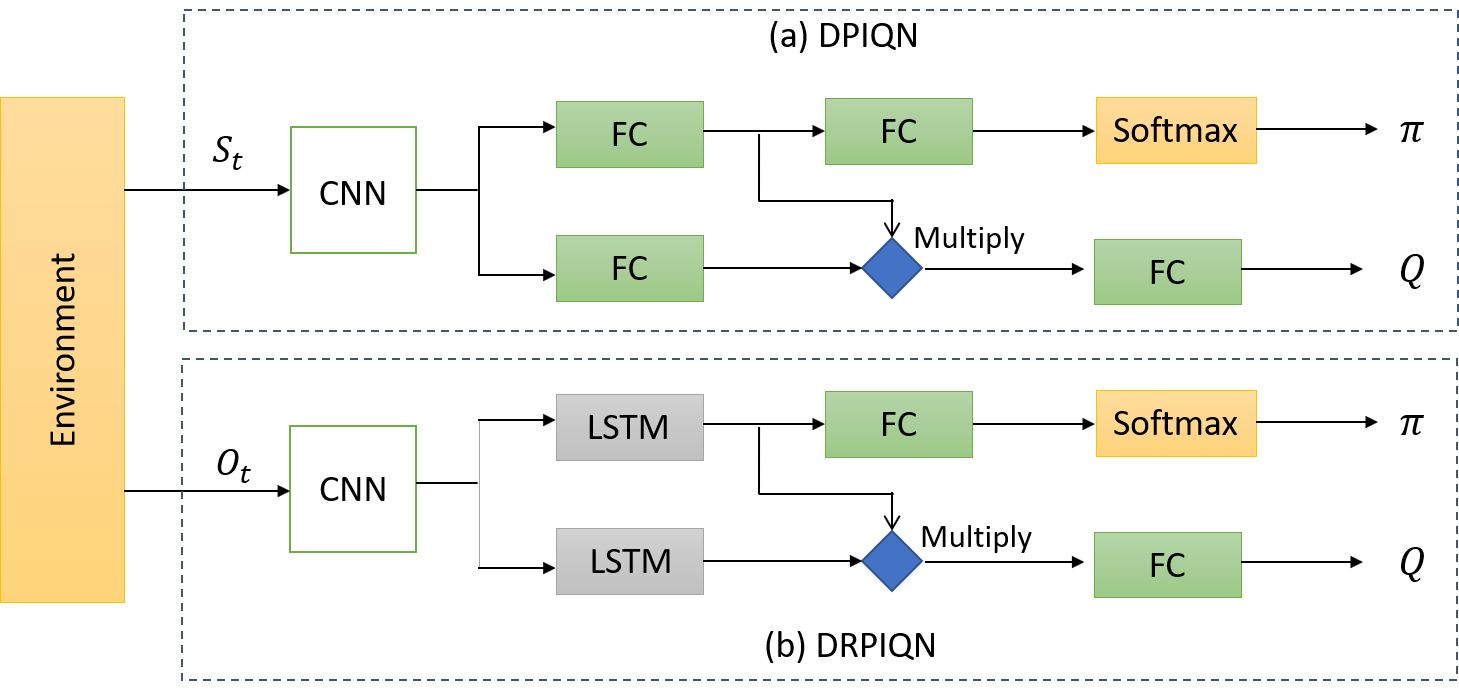}
\caption{Architecture of DPIQN and DRPIQN.}
\label{fig_DPIQN_DRPIQN} 
\end{figure}

Scaling to a system of many agents in partial observable domains is a challenging problem. Gupta et al. \cite{Gupta2017} extended the \emph{curriculum learning} technique to an MAS, which integrates with three classes of deep RL, including policy gradient, temporal-difference error, and actor-critic methods. The curriculum principle is to start learning to complete simple tasks first to accumulate knowledge before proceeding to perform complicated tasks. This is suitable with an MAS environment where fewer agents initially collaborate before extending to accommodate more and more agents to complete increasingly difficult tasks. Experimental results show the vitality of the curriculum learning method in scaling deep RL algorithms to complex multi-agent problems. 

Hong et al. \cite{Hong2018} introduced a \emph{deep policy inference Q-network} (DPIQN) to model multi-agent systems and its enhanced version \emph{deep recurrent policy inference Q-network} (DRPIQN) to cope with partial observability. Both DPIQN and DRPIQN are learned by adapting network's attention to policy features and their own Q-values at various stages of the training process (Fig. \ref{fig_DPIQN_DRPIQN}). Experiments show the better overall performance of both DPIQN and DRPIQN over the baseline DQN and DRQN \cite{Hausknecht2015}. Also in the context of partial observability, but extended to multi-task, multi-agent problems, Omidshafiei et al. \cite{Omidshafiei2017} proposed a method called \emph{multi-task multi-agent RL} (MT-MARL) that integrates hysteretic learners \cite{Matignon2007}, DRQNs \cite{Hausknecht2015}, distillation \cite{Rusu2015}, and concurrent experience replay trajectories (CERTs), which are a decentralized extension of experience replay strategy proposed in \cite{Mnih2015}. The agents are not explicitly provided with task identity (thus partial observability) whilst they cooperatively learn to complete a set of \emph{decentralized POMDP} tasks with sparse rewards. This method however has a disadvantage that cannot perform in an environment with heterogeneous agents.

Apart from partial observability, there are circumstances that agents must deal with extremely noisy observations, which are weakly correlated with the true state of the environment. Kilinc and Montana \cite{Kilinc2018} introduced a method denoted as MADDPG-M that combines DDPG and a communication medium to address these circumstances. Agents need to decide whether their observations are informative to share with other agents and the communication policies are learned concurrently with the main policies through experience. Recently, Foerster et al. \cite{Foerster2018a} proposed \emph{Bayesian action decoder} (BAD) algorithm for learning multiple agents with cooperative partial observable settings. A new concept, namely public belief MDP, is introduced based on BAD that employs an approximate Bayesian update to attain a public belief with publicly observable features in the environment. BAD relies on a factorised and approximate belief state to discover conventions to enable agents to learn optimal policies efficiently. This is closely relevant to \emph{theory of mind} that humans normally use to interpret others' actions. Experimental results on a proof-of-principle two-step matrix game and the cooperative partial-information card game Hanabi demonstrate the efficiency and superiority of the proposed method against traditional policy gradient algorithms. 

\subsubsection{MAS training schemes}

The direct extension of single agent deep RL to multi-agent environment is to learn each agent independently by considering other agents as part of the environment as the independent Q-learning algorithm proposed in \cite{Tampuu2017}. This method is vulnerable to overfitting \cite{Lanctot2017} and computationally expensive, and therefore the number of agents involved is limited. An alternative and popular approach is the \emph{centralized learning and decentralized execution} where a group of agents can be trained simultaneously by applying a centralized method via an open communication channel \cite{Kraemer2016}. Decentralized policies where each agent can take actions based on its local observations have an advantage under partial observability and in limited communications during execution. Centralized learning of decentralized policies has become a standard paradigm in multi-agent settings because the learning process may happen in a simulator and a laboratory where there are no communication constraints, and extra state information is available \cite{Kraemer2016, Foerster2018b}.

Gupta et al. \cite{Gupta2017} examined three different training schemes for a MAS, which consists of centralized learning, concurrent learning and parameter sharing. Centralized policy attempts to obtain a joint action from joint observations of all agents whilst the concurrent learning trains agents simultaneously using the joint reward signal. In the latter, each agent learns its own policy independently based on private observation. Alternatively, the parameter sharing scheme allows agents to be trained simultaneously using the experiences of all agents although each agent can obtain unique observations. With the ability to execute decentralized policies, parameter sharing can be used to extend a single agent deep RL algorithm to accommodate a system of many agents. Particularly, the combination of \emph{parameter sharing and TRPO}, namely PS-TRPO, has been proposed in \cite{Gupta2017}, and briefly summarized in Algorithm \ref{alg1}. The PS-TRPO has demonstrated great performance when dealing with high-dimensional observations and continuous action spaces under partial observability.

%Algorithm 1: PS-TRPO
%	Initialize parameters of policy network $\theta_0$ and trust region size $\varDelta$
%	for i = 0, 1, …, do
%	Generate trajectories for all agents as τ ~ π using the policy with shared parameters.
%	For each agent m, compute the advantage value Aπ(o,m,a) where m is the agent index.
%	Search π that maximizes L(0) subject to DKL where DKL is the KL divergence between two policy distributions, and pthetak are the discounted state-visitation frequencies induced by pi. 

\begin{algorithm}[!t]
\caption{PS-TRPO}
\label{alg1}
\begin{algorithmic}[1]
\State Initialize parameters of policy network $\Theta_0$, and trust region size $\Delta$
\For{$i  \longleftarrow 0,1, ...$}
\State Generate trajectories for all agents as $\tau \sim \pi_{\theta_i}$ using the policy with shared parameters.
\State For each agent $m$, compute the advantage values $A_{\pi_{\theta_i}} (o^m, m, a^m)$ with $m$ is the agent index.
\State Search $\pi_{\theta_{i+1}}$ that maximizes $L(\theta) = E_{o \sim p_{\theta_k}, a \sim \pi_{\theta_k}} \left[  \frac{\pi_{\theta}(a | o, m)} {\pi_{\theta_k}(a | o,m)} A_{\theta_k}(o, m, a) \right]$
 \newline \hspace*{0.7cm} subject to $\bar{D}_{KL}(\pi_{\theta_i} \parallel \pi_{\theta_{i+1}}) \leq \Delta $ where $D_{KL}$ is the KL divergence between distributions of  
 \newline \hspace*{0.7cm} two policies, and $p_\theta$ are the discounted frequencies of state visitation caused by $\pi_{\theta}$.
\EndFor
\State \textbf{end for}
\end{algorithmic}
\end{algorithm}

%\begin{algorithm}[H]
% \caption{PS-TRPO}
% \KwData{this text}
% \KwResult{how to write algorithm with \LaTeX2e }
% initialization\;
% \While{not at end of this document}{
%  read current\;
%  \eIf{understand}{
%   go to next section\;
%   current section becomes this one\;
%   }{
%   go back to the beginning of current section\;
%  }
% }
%\end{algorithm}

Foerster et al. \cite{Foerster2016a} introduced \emph{reinforced inter-agent learning} (RIAL) and \emph{differentiable inter-agent learning} (DIAL) methods based on the centralized learning approach to improve agent learning communication. In RIAL, deep Q-learning has a recurrent structure to address the partial observability issue, in which independent Q-learning offers individual agents to learn their own network parameters. DIAL pushes gradients from one agent to another through a channel, allowing end-to-end backpropagation across agents. Likewise, Sukhbaatar et al. \cite{Sukhbaatar2016} developed communication neural net (CommNet) allowing dynamic agents to learn continuous communication alongside their policy for fully cooperative tasks. Unlike CommNet, He et al. \cite{He2016} proposed a method namely \emph{deep reinforcement opponent network} (DRON) that encodes observation of the opponent agent into DQN to jointly learn a policy and behaviors of opponents without domain knowledge.

Kong et al. \cite{Kong2017a, Kong2017b} incorporated both decentralized and centralized perspectives into the hierarchical master-slave architecture to form a model named \emph{master-slave multi-agent RL} (MS-MARL) to solve the communication problem in MAS. The master agent receives and collectively processes messages from slave agents and then generates unique instructive messages to each slave agent. Slave agents use their own information and instructive messages from the master agent to take actions. This model significantly reduces the communication burden within a MAS compared to the peer-peer architecture, especially when the system has many agents.

\begin{figure}[!h]
\centering
\includegraphics[width=0.8\columnwidth]{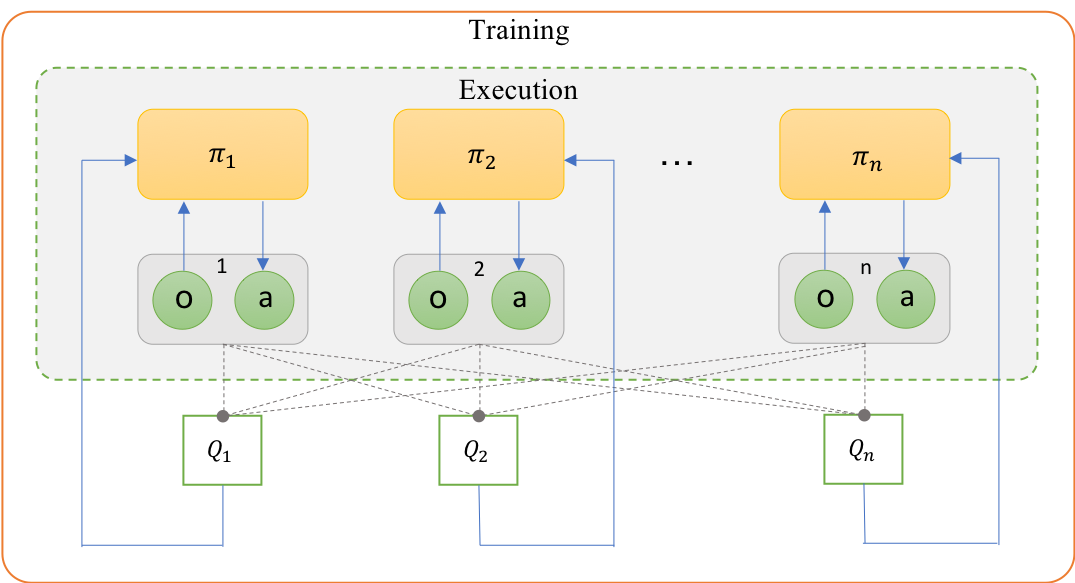}
\caption{Centralized learning and decentralized execution based MADDPG where policies of agents are learned by the centralized critic with augmented information from other agents' observations and actions.}
\label{fig_maddpg} 
\end{figure} 

Lowe et al. \cite{Lowe2017} proposed \emph{multi-agent deep deterministic policy gradient} (MADDPG) method based on the actor-critic policy gradient algorithms. MADDPG features the centralized learning and decentralized execution paradigm in which the critic uses extra information to ease training process whilst actors take actions based on their own local observations. Fig. \ref{fig_maddpg} illustrates the multi-agent decentralized actor and centralized critic components of MADDPG where only actors are used during the execution phase.

Recently, Foerster et al. \cite{Foerster2018b} introduced another multi-agent actor-critic method, namely counterfactual multi-agent (COMA), which was also relied on the centralized learning and decentralized execution scheme. Unlike MADDPG \cite{Lowe2017}, COMA can handle the \emph{multi-agent credit assignment} problem \cite{Harati2007} where agents are difficult to work out their contribution to the team's success from global rewards generated by joint actions in cooperative settings. COMA however has a disadvantage that focuses only on discrete action space whilst MADDPG is able to learn continuous policies effectively.

\subsubsection{Continuous action spaces}
Most deep RL models can only be applied to discrete spaces \cite{Lillicrap2015}. For example, DQN \cite{Mnih2015} is limited only to problems with discrete and low-dimensional action spaces, although it can handle high-dimensional observation spaces. DQN aims to find action that has maximum action-value, and therefore requires an iterative optimization process at every step in the continuous action (state) spaces. Discretising the action space is a possible solution to adapt deep RL methods to continuous domains. However, this creates many problems, notably is the curse of dimensionality: the exponential increase of action numbers against the number of degrees of freedom. 

Schulman et al. \cite{Schulman2015} proposed \emph{trust region policy optimization} (TRPO) method, which can be extended to continuous states and actions, for optimizing stochastic control policies in the domain of robotic locomotion and image-based game playing. Lillicrap et al. \cite{Lillicrap2015} introduced an off-policy algorithm, namely \emph{deep deterministic policy gradient} (DDPG), which utilizes the actor-critic architecture \cite{Konda2000, Mnih2016} to handle the continuous action spaces. Based on the deterministic policy gradient (DPG) \cite{Silver2014}, DDPG deterministically maps states to specific actions using a parameterized actor function while keeping DQN learning on the critic side. This approach however requires a large number of training episodes to find solutions, as found common in model-free reinforcement methods. Heess et al. \cite{Heess2015} extended DDPG to \emph{recurrent DPG} (RDPG) to handle problems with continuous action spaces under partial observability, where the true state is not available to the agents when making decisions. Recently, Gupta et al. \cite{Gupta2017} introduced the PS-TRPO method for multi-agent learning (see Algorithm \ref{alg1}). This method is based on the foundation of TRPO so that it can deal with continuous action spaces effectively.  

\subsubsection{Transfer Learning for MADRL}
Training the Q-network or generally a deep RL model of a single agent is often very computationally expensive. This problem is significantly severe for a system of multiple agents. To improve the performance and reduce computational costs during training process of multiple deep RL models, several studies have promoted transfer learning for deep RL.

Rusu et al. \cite{Rusu2015, Rusu2016} proposed \emph{policy distillation} method and \emph{progressive neural networks} to promote transfer learning in the context of deep RL. These methods however are computationally complex and expensive \cite{Kirkpatrick2017}. Yin and Pan \cite{Yin2017} likewise introduced another policy distillation architecture to apply knowledge transfer for deep RL. That method reduces training time and outperforms DQNs but its exploration strategy is still not efficient. Parisotto et al. \cite{Parisotto2015} proposed the \emph{actor-mimic} method for multi-task and transfer learning that improves learning speed of a deep policy network. The network can obtain an expert performance on many games simultaneously, although its model is not so complex. That method however requires a sufficient level of similarity between source and target tasks and is vulnerable to negative transfer. 

Egorov \cite{Egorov2016} reformulated a multi-agent environment into an image like representation and utilized CNNs to estimate Q-values for each agent in question. That approach can address the scalability problem in MAS when the transfer learning method can be used to speed up the training process. A policy network trained on a different but related environment is used for learning process of other agents to reduce computational expenses. Experiments carried out on the pursuit-evasion problem \cite{Chung2011} show the effectiveness of the transfer learning approach in the multi-agent domain. 

Table \ref{table_sum} presents a summary of reviewed papers that address different multi-agent learning challenges. It can be seen that many extensions of DQN have been proposed in the literature whilst policy-based or actor-critic methods have not adequately been explored in multi-agent environments. 

\begin{scriptsize}
\begin{longtable}{| p{0.1\textwidth}| p{0.4\textwidth}| p{0.19\textwidth}| p{0.19\textwidth}|}
\caption{Multi-agent learning challenges and their solving methods}
\label{table_sum}\\
\hline
\hline
Challenges & Value-based & Actor-critic & Policy-based\\
\hline
Partial observability
&DRQN \cite{Hausknecht2015}; DDRQN \cite{Foerster2016b}; RIAL and DIAL \cite{Foerster2016a}; Action-specific DRQN \cite{Zhu2017}; MT-MARL \cite{Omidshafiei2017}; PS-DQN \cite{Gupta2017}; RL as a Rehearsal (RLaR) \cite{Kraemer2016}
&PS-DDPG and PS-A3C \cite{Gupta2017}; MADDPG-M \cite{Kilinc2018}
&DPIQN and DRPIQN \cite{Hong2018}; PS-TRPO \cite{Gupta2017}; Bayesian action decoder (BAD) \cite{Foerster2018a}\\
\hline
Non-stationarity
&DRUQN and DLCQN \cite{Castaneda2016}; Multi-agent concurrent DQN \cite{Diallo2017}; Recurrent DQN-based multi-agent importance sampling and fingerprints \cite{Foerster2017}; Hysteretic-DQN \cite{Omidshafiei2017}; Lenient-DQN \cite{Palmer2018}; WDDQN \cite{Zheng2018}
&MADDPG \cite{Lowe2017}; PS-A3C \cite{Gupta2017}
&PS-TRPO \cite{Gupta2017}\\
\hline
Continuous action spaces
& \,
&Recurrent DPG \cite{Heess2015}; DDPG \cite{Lillicrap2015}
&TRPO \cite{Schulman2015}; PS-TRPO \cite{Gupta2017}\\
\hline
Multi-agent training schemes
&Multi-agent extension of DQN \cite{Tampuu2017}; RIAL and DIAL \cite{Foerster2016a}; CommNet \cite{Sukhbaatar2016}; DRON \cite{He2016}; MS-MARL \cite{Kong2017a, Kong2017b}; Linearly fuzzified joint Q-function for MAS \cite{Luviano-Cruz2018}
&MADDPG \cite{Lowe2017}; COMA \cite{Foerster2018b}
&\,\\
\hline
Transfer learning in MAS
&Policy distillation \cite{Rusu2015}; Multi-task policy distillation \cite{Yin2017}; Multi-agent DQN \cite{Egorov2016}
&Progressive networks \cite{Rusu2016}
&Actor-Mimic \cite{Parisotto2015}\\
\hline
\hline
\end{longtable}
\end{scriptsize}

\subsection{MADRL Applications}
Since the success of deep RL marked by the DQN proposed in \cite{Mnih2015}, many algorithms have been proposed to integrate deep learning to multi-agent learning. These algorithms can solve complex problems in various fields. This section provides a survey of these applications with a focus on the integration of deep learning and MARL. Table \ref{table_app_sum} summarizes the features and limitations of approaches to these applications.

%\begin{table}[!t]
%\centering
%\resizebox{\textwidth}{!}{%
%\begin{longtable}{|p{2cm}|p{2cm}|p{5cm}|p{4cm}|}
\afterpage{
\begin{scriptsize}
\begin{longtable}{| p{0.09\textwidth}| p{0.09\textwidth}| p{0.4\textwidth}| p{0.32\textwidth}|}
\caption{A summary of typical MADRL applications in different fields}
\label{table_app_sum}\\
\hline
\hline
Applications & Basic DLR & Features & Limitations\\
\hline
Federated control \cite{Kumar2017} 
&Hierarchical-DQN (h-DQN) \cite{Kulkarni2016}
&$\bullet$Divide the control problem into disjoint subtasks and leverage \emph{temporal abstractions}.\newline
$\bullet$Use \emph{meta-controller} to guide decentralized controllers.\newline
$\bullet$Able to solve distributed scheduling problems such as multi-task dialogue, urban traffic control.
&$\bullet$Does not address the non-stationarity problem.\newline
$\bullet$Number of agents is currently limited at six.\newline
$\bullet$Meta-controller's optimal policy becomes complicated and inefficient when the number of agents increases.\\
\hline
Large-scale fleet management \cite{Lin2018}
&Actor-critic and DQN
&$\bullet$Reallocate vehicles ahead of time to balance the transport demands and supplies.\newline
$\bullet$Geographic context and collaborative context are integrated to coordinate agents.\newline
$\bullet$Two proposed algorithms, \emph{contextual multi-agent actor-critic} and \emph{contextual deep Q-learning}, can achieve explicit coordination among thousands of agents.
&$\bullet$Can only deal with discrete actions and each agent has a small (simplified) action space.\newline
$\bullet$Assume that agents in the same region at the same time interval (i.e. same spatial-temporal state) are homogeneous.\\
\hline
Swarm systems \cite{Huttenrauch2017}
&DQN and DDPG
&$\bullet$Agents can only observe local environment (partial observability) but not the global state.\newline
$\bullet$Use guided approach for multi-agent learning where actors make decisions based on locally sensed information whilst critic has central access to global state.
&$\bullet$Can only work with homogeneous agents.\newline
$\bullet$Unable to converge to meaningful policies in huge dimensionality and partial observed problem.\\
\hline
Traffic lights control \cite{Calvo2018}
&DDDQN and IDQN
&$\bullet$Learning multiple agents is performed using IDQN where each agent is modelled by DDDQN.\newline
$\bullet$First approach to address heterogeneous multi-agent learning in urban traffic control.\newline
$\bullet$Fingerprint technique is used to stabilize the experience replay memory to handle non-stationarity.
&$\bullet$The proposed deep RL approach learns ineffectively in high traffic conditions.\newline
$\bullet$The fingerprint does not improve the performance of experience replay although the latter is required for efficient learning.\\
\hline
Task and resources allocation \cite{Noureddine2017}
&CommNet \cite{Sukhbaatar2016}
&$\bullet$Propose distributed task allocation where agents can request help from cooperating neighbors.\newline
$\bullet$Three types of agents are defined: manager, participant and mediator.\newline
$\bullet$Communication protocol is learned simultaneously with agents' policies through CommNet.
&$\bullet$May not be able to deal with heterogeneous agents.\newline
$\bullet$Computational deficiencies regarding the decentralization and reallocation characteristics.\newline
$\bullet$Experiments only on small state action spaces.\\
\hline
Energy sharing optimization \cite{Prasad2018}
&DQN
&$\bullet$Each building is characterized by a DRL agent to learn appropriate actions independently.\newline
$\bullet$Agents' actions include: consume and store excess energy, request neighbor or supply grid for additional energy, grant or deny requests.\newline
$\bullet$Agents collaborate via shared or global rewards to achieve a common goal, i.e. zero-energy status.
&$\bullet$Agents' behaviors cannot be observed in an online fashion.\newline
$\bullet$Limited number of houses, currently ten houses at maximum were experimented.\newline
$\bullet$Energy price is not considered.\\
\hline
Keepaway soccer \cite{Kurek2016}
&DQN
&$\bullet$Low-dimensional state space, described by only 13 variables.\newline
$\bullet$Heterogeneous MAS, each agent has different experience replay memory and different network policy.
&$\bullet$Number of agents is limited, currently setting with 3 keepers vs. 2 takers.\newline
$\bullet$Heterogeneous learning speed is significantly lower than homogeneous case.\\
\hline
Action markets \cite{Schmid2018}
&DQN
&$\bullet$Agents can trade their atomic actions in exchange for environmental reward.\newline
$\bullet$Reduce greedy behavior and thus negative effects of individual reward maximization.\newline
$\bullet$The proposed approach significantly increases the overall reward compared to methods without action trading.  
&$\bullet$Agents cannot find prices for actions by themselves because they are given at design time.\newline
$\bullet$Strongly assume that agents cannot cheat on each other by making offers which they do not hold afterwards.\\
\hline
Sequential social dilemma (SSD) \cite{Leibo2017}
&DQN
&$\bullet$Introduce an SSD model to extend MGSD to capture sequential structure of real-world social dilemmas.\newline
$\bullet$Describe SSDs as general-sum Markov games with partial observations.\newline
$\bullet$Multi-agent DQN is used to find equilibria of SSD problems.
&$\bullet$Assume agent's learning is independent and regard the others as part of the environment.\newline
$\bullet$Agents do not recursively reason about one another's learning.\\
\hline
Common-pool resource (CPR) appropriation \cite{Perolat2017}
&DQN
&$\bullet$Introduce a new CPR appropriation model using an MAS containing spatially and temporally environment dynamics.\newline
$\bullet$Use the \emph{descriptive agenda} \cite{Shoham2007} to describe the behaviors emerging when agents learn in the presence of other learning agents.\newline
$\bullet$Simulate multiple independent agents with each learned by DQN.
&$\bullet$Single agent DQN is extended to multi-agent environment where the Markov assumption is no longer hold.\newline
$\bullet$Agents do not do anything of \emph{rational negotiation}, e.g. bargaining, building consensus, or making appeals.\\
\hline
\hline
%\label{table_app_sum}
\end{longtable}
%}
%\end{table}
\end{scriptsize}
}

Prasad and Dusparic \cite{Prasad2018} introduced a MADRL model to deal with energy sharing problem in a zero-energy community that comprises a collection of zero energy buildings, which have the total energy use over a year smaller than or equal to the renewables generation within each building. A deep RL agent is used to characterize each building to learn appropriate actions in sharing energy with other buildings. The global reward is modelled by the negative of the community energy status as $reward=-(\sum_{i=1}^{n}c(h_i)-g(h_i))$, where $c(h_i )$ and $g(h_i)$ are the energy consumed and energy generated by the $i$th building. The community monitoring service is introduced to manage the group membership activities such as joining and leaving the group or maintaining a list of active agents. Experiments show the superiority of the proposed model compared to the random action selection strategy in terms of net zero energy balance as a community. A limitation of the proposed approach lies in the episodic learning manner so that agent's behaviors cannot be observed in an online fashion. Further drawbacks include that the current experiments were not performed on a large scale, i.e. ten houses at maximum, and energy price scheme is not considered.

\begin{figure}[!h]
\centering
\includegraphics[width=0.9\columnwidth]{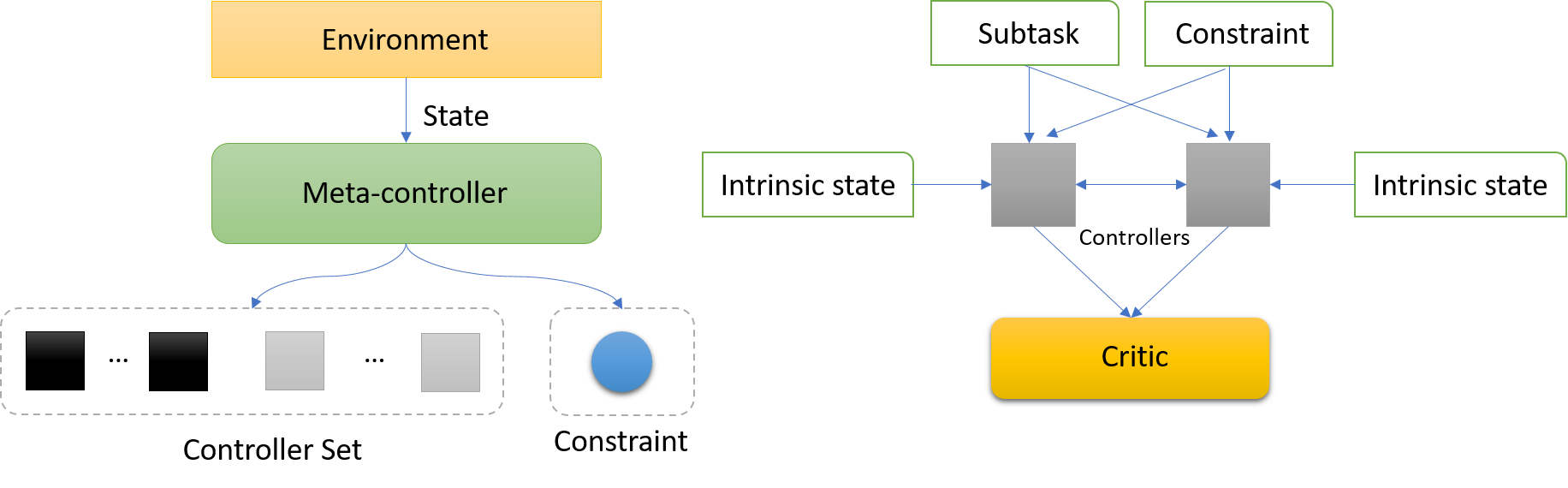}
\caption{Federated control with hierarchical MADRL method.}
\label{fig_federated_control} 
\end{figure} 

A combination of hierarchical RL and MADRL methods was developed to coordinate and control multiple agents in problems where agents' privacy is prioritized \cite{Kumar2017}. Such distributed scheduling problems could be a multi-task dialogue where an automated assistant needs to help a user to plan for several independent tasks, for example, purchase a train ticket to the city, book a movie ticket, and make a dinner reservation in a restaurant. Each of these tasks is handled by a decentralized controller whilst the assistant is a meta-controller which benefits from \emph{temporal abstractions} to lessen the communication complexity, and thus able to find a globally consistent solution for the user (Fig. \ref{fig_federated_control}). On the other hand, Leibo et al. \cite{Leibo2017} introduced a \emph{sequential social dilemma} (SSD) model based on general-sum Markov games under partial observability to address the evolution of cooperation in MAS. Being able to capture sequential structure of real-world social dilemmas, SSD is an extension of \emph{matrix game social dilemma} (MGSD) that has been applied to various phenomena in social science and biology \cite{deCote2006, Bloembergen2015, Yu2015}. The general-sum modelling requires solving algorithms to either track different potential equilibria for each agent or be able to find cyclic strategy consisting of multiple policies learned by using different state space sweeps \cite{Zinkevich2006, Perolat2016}. DQN is utilized to characterize self-interested independent learning agents to find equilibria of the SSD, which cannot be solved by the standard evolution and learning methods used for MGSD \cite{Kleiman-Weiner2016}. Perolat et al. \cite{Perolat2017} also demonstrated the application of MADRL in social science phenomenon, i.e. the common-pool resource (CPR) appropriation. The proposed method comprises a spatially and temporally dynamic CPR environment \cite{Janssen2010} and an MAS of independent self-interested DQNs. Through the RL trial and error process, the CPR appropriation problem is solved by self-organization that adjusts the incentives felt by independent individual agents over time.

Huttenrauch et al. \cite{Huttenrauch2017} formulated swarm systems as a special case of the decentralized POMDP \cite{Oliehoek2012} and used an actor-critic deep RL approach to control a group of cooperative agents. The Q-function is learned using a global state information, which can be the view of a camera capturing the scene in swarm robotics examples. The group can perform complicated tasks such as search and rescue or distributed assembly although individual agent has limited sensory capability. That model has a drawback as it assumes agents to be homogenous. Calvo and Dusparic \cite{Calvo2018} proposed the use of IDQN to address the heterogeneity problem in multi-agent environment of urban traffic light control. Each agent is learned by the \emph{dueling double deep Q-network} (DDDQN), which integrates dueling networks, double DQN and prioritized experience replay. Heterogenous agents are trained independently and simultaneously, considering other agents as part of the environment. The non-stationarity of the multi-agent environment is dealt with by a technique of fingerprinting that disambiguates the age of training samples and stabilizes the replay memory. 

A special application of DQN to the heterogeneous MAS where the state space is low-dimensional was presented in \cite{Kurek2016}. Experiments are performed on a multi-agent Keepaway soccer problem whose state comprises only 13 variables. To handle heterogeneity, each DQN agent is set up with different experience replay memory and different neural network. Agents cannot communicate with each other but only can observe others' behaviors. While DQNs can enhance results in terms of game score in the heterogeneous team learning settings in low-dimensional environments, their learning process is significantly slower than those in the homogeneous cases. 

Establishing communication channels among agents during learning is an important step in designing and constructing MADRL algorithms. Nguyen et al. \cite{Nguyen2018} characterized the communication channel via human knowledge represented by images and allow deep RL agents to communicate using these shared images. The \emph{asynchronous advantage actor-critic} (A3C) algorithm \cite{Mnih2016} is used to learn optimal policy for each agent, which can be extended to multiple heterogeneous agents. On the other hand, Noureddine et al. \cite{Noureddine2017} introduced a method, namely task allocation process using cooperative deep RL, to allow multiple agents to interact with each other and allocate resources and tasks effectively. Agents can request help from their cooperative neighbors in a loosely coupled distributed multi-agent environment. The CommNet model \cite{Sukhbaatar2016} is used to facilitate communications among agents, which are characterized by DRQN \cite{Hausknecht2015}. Experimental results demonstrate the great capability of that method in handling complicated task allocation problem. One of the drawbacks of that algorithm however is its limited ability in managing heterogeneous agents. In addition, its decentralization and reallocation characteristics also pose disadvantages in terms of computational time and communication overhead.

\begin{figure}[!h]
\centering
\includegraphics[width=0.9\columnwidth]{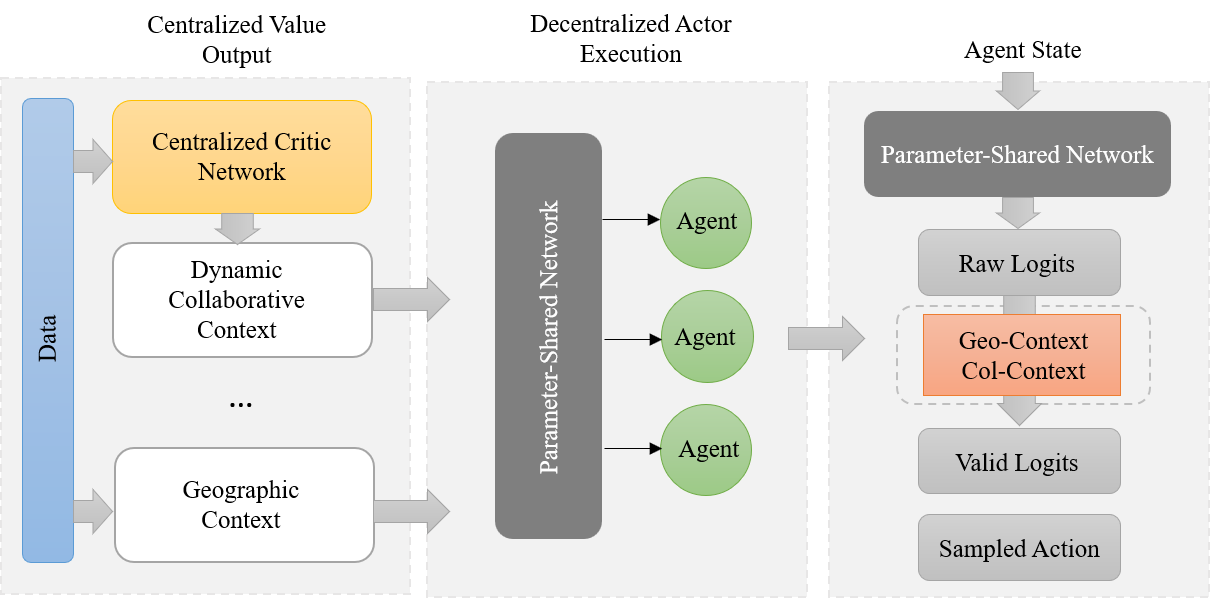}
\caption{The contextual multi-agent actor-critic architecture proposed in \cite{Lin2018} where decentralized execution is coordinated using the centralized value network's outputs as illustrated in the left part whilst the right part shows how context is embedded to the policy network.}
\label{fig_contextual_ac} 
\end{figure}

Lin et al. \cite{Lin2018} addressed the large-scale fleet management using MADRL by proposing two algorithms, namely \emph{contextual deep Q-learning} and \emph{contextual multi-agent actor-critic}. These algorithms aim to balance the difference between demand and supply by reallocating transportation resources that help to reduce traffic congestion and increase transportation efficiency. The contextual multi-agent actor-critic model is illustrated in Fig. \ref{fig_contextual_ac} where a parameter-shared policy network is used to coordinate the agents, which represent available vehicles or equivalently the idle drivers.

Recently, Schmid et al. \cite{Schmid2018} introduced an interesting approach to an MAS where agents can trade their actions in exchange for other resources, e.g. environmental rewards. The action trading was inspired by the fundamental theorem of welfare economics that competitive markets adjust towards the Pareto efficiency. Specifically, agents need to extend their action spaces and learn two policies simultaneously: one for the original stochastic reward and another for trading environmental reward. The behavior market realized from the action trading helps mitigate greedy behavior (like the tit-for-tat game theoretic strategy proposed in \cite{Lerer2017}), enable agents to incentivize other agents and reduce the negative effects of individual reward maximization. Simulation results on the iterated matrix game and the Coin game show the effectiveness of the action trading method as it increases the social welfare, measured in terms of overall rewards of all agents.

\section{Conclusions and Research Directions}
This paper presents an overview of different challenges in multi-agent learning and solutions to these challenges using deep RL methods. We group the surveyed papers into five categories, including non-stationarity, partial observability, multi-agent training schemes, multi-agent transfer learning, and continuous state and action spaces. We have highlighted advantages and disadvantages of the approaches to address the challenges. Applications of MADRL methods in different fields are also reviewed thoroughly. We have found that the integration of deep learning into traditional MARL methods has been able to solve many complicated problems, such as urban traffic light control, energy sharing problem in a zero-energy community, large-scale fleet management, task and resources allocation, swarm robotics, and social science phenomena. The results indicate that deep RL-based methods provide a viable approach to handling complicated tasks in the MAS domain.

Learning from demonstration including \emph{imitation learning} and \emph{inverse RL} has demonstrated effectiveness in single agent deep RL \cite{Piot2017}. On one hand, imitation learning tries to map between states to actions as a supervised approach. It directly generalizes the expert strategy to unvisited states so that it is close to a multi-class classification problem in cases of finite action set. On the other hand, inverse RL agent needs to infer a reward function from the expert demonstrations. Inverse RL assumes that the expert policy is optimal regarding the unknown reward function \cite{Hadfield-Menell2016, Hadfield-Menell2017}. These methods however have not yet been explored fully in multi-agent environments. Both imitation learning and inverse RL have great potential for applications in MAS. It is expected that they can reduce the learning time and improve the effectiveness of MAS. A very straightforward challenge arose from these applications is the requirement of multiple experts who are able to demonstrate the tasks collaboratively. Furthermore, the communication and reasoning capabilities of experts are difficult to be characterized and modelled by autonomous agents in the MAS domain. These pose important research questions towards extensions of imitation learning and inverse RL to MADRL methods. In addition, for complicated tasks or behaviors which are difficult for humans to demonstrate, there is a need of alternative methods that allow human preferences to be integrated into deep RL \cite{Christiano2017, Nguyen2018b, Nguyen2018}.

Deep RL has considerably facilitated autonomy, which allows to deploy many applications in robotics or autonomous vehicles. The most common drawback of deep RL models however is the ability to interact with human through human-machine teaming technologies. In complex and adversarial environments, there is a critical need for human intellect teamed with technology because humans alone cannot sustain the volume, and machines alone cannot issue creative responses when new situations are introduced. Recent advances of \emph{human-on-the-loop} architecture \cite{Nahavandi2017} can be fused with MADRL to integrate humans and autonomous agents to deal with complex problems. In the conventional \emph{human-in-the-loop} setting, agents perform their assigned tasks autonomously for a period, then stop and wait for human commands before continuing in this rate-limited fashion. In human-on-the-loop, agents execute their tasks autonomously until completion, with a human in a monitoring or supervisory role reserving the ability to intervene in operations carried out by agents. A human-on-the-loop-based architecture can be fully autonomous if human supervisors allow task completion by agents entirely on their own \cite{Nahavandi2017}.

Model-free deep RL has been able to solve many complicated problems both in single agent and multi-agent domains. This category of methods however requires a huge number of samples and long learning time to achieve a good performance. \emph{Model-based} methods have demonstrated effectiveness in terms of sample efficiency, transferability and generality in various problems using single as well as multi-agent models. Although the deep learning extensions of the model-based methods have been studied recently in single agent domain, e.g. \cite{Levine2016, Gu2016, Finn2017, Nagabandi2018, Serban2018, Corneil2018}, these extensions have not been investigated widely in the multi-agent domain. This creates a research gap that could be developed to a research direction in model-based MADRL. In addition, dealing with high-dimensional observations using model-based approaches or combining elements of model-based planning and model-free policy is another active, exciting but under-explored research area. 

Scaling to large systems, especially dealing with many heterogeneous agents, has been a primary challenge in RL research domain since its first days. As the world dynamics become more and more complex, this challenge has always been required to resolve. Since agents have common behaviors such as actions, domain knowledge, and goals (homogeneous agents), the scalability can be achievable by (partially) centralized training and decentralized execution \cite{Rashid2018, Foerster2018c}. In the heterogeneous setting with many agents, the key challenge is how to provide the most optimal solution and maximize the task completion success based on self-learning with effective coordinative and cooperative strategies among the agents. This is more problematic in hostile environments where communications among agents are limited and in scenarios that involve more heterogeneous agents as the credit assignment problem become increasingly difficult. A research direction to address these difficulties is well worth an investigation. 

Regarding applications of multi-agent learning, there have been many studies using traditional MARL methods to solve various problems such as controlling a group of autonomous vehicles or drones \cite{Hung2017}, robot soccer \cite{Schwab2018}, controlling traffic signals \cite{Mousavi2017}, coordinating collaborative bots in factories and warehouse \cite{Kattepur2018}, controlling electrical power networks \cite{Rahman2015} or optimizing distributed sensor networks \cite{He2015}, automated trading \cite{Pendharkar2018}, machine bidding in competitive e-commerce and financial markets \cite{Brandouy2011}, resource management \cite{Hussin2015}, transportation \cite{Fernandez-Gauna2015}, and phenomena of social sciences \cite{Leibo2017}. Since the emergence of DQN \cite{Mnih2015}, efforts to extend traditional RL to deep RL in the multi-agent domain have been found in the literature but they are still very limited (see Table \ref{table_app_sum} for applications available in the current literature). Many applications of MARL can now be solved effectively by MADRL based on its high-dimension handling capability. Therefore, there is a need of further empirical research to apply MADRL methods to effectively solve complex real-world problems such as the aforementioned applications.

\end{document}